\documentclass{article}
\usepackage[final]{corl_2026}  
\usepackage{iftex}
\ifXeTeX  
  \usepackage{fontspec}
\fi
\usepackage{amsmath,amssymb}
\usepackage{graphicx}
\usepackage{booktabs}
\usepackage{multirow}  
\usepackage{placeins}  

\newcommand{\DET}{\textsc{det}}
\newcommand{\NODET}{\textsc{no\_det}}

\title{What World Model Does Active Object Search Need?\\ A Study under Imperfect Detection}

\author{
  Kyeongryeol Go\\
  Superb AI\\
  \texttt{krgo@superb-ai.com}
}

\begin{document}
\maketitle

\begin{abstract}
In active object search, a robot must decide where to look and when to stop. Imperfect detection complicates both decisions: a non-detection may indicate occlusion, an out-of-view target, or a false negative, while a detection may be a false positive. This raises the question of which representations a world model needs to support reliable search under imperfect detection. We introduce Bayesian Active Miss-aware Object Search (BAMOS) to investigate how target belief, occupancy belief, and detector errors affect search decisions. Through comprehensive ablations in Isaac Sim and HSSD houses, we examine the roles of these representations and their update mechanisms across different detector configurations. Notably, target belief and its update mechanisms are critical to search performance, with their contribution generally increasing as detection errors become more prevalent. These findings highlight the importance of maintaining and updating target beliefs to support reliable active search under uncertain detector observations, providing insights into the design of world models for embodied agents. Code is available in our \href{https://github.com/gokyeongryeol/BAMOS}{\textcolor{magenta}{GitHub repository}}.
\end{abstract}


\section{Introduction}

In \emph{active object search}, a robot chooses where to point its camera until it can state where a target object
is~\citep{aydemir2013active,zheng2021multi}. Consider a mobile manipulator instructed to ``bring me the mug'' while the mug is
behind a box on a shelf. Before grasping, it must decide where to look next and when it has gathered enough evidence. When a
view yields no detection, the target may be occluded, outside the frame, or visible but missed by the detector (a
\emph{false negative}). Open-vocabulary detectors miss targets often enough that a non-detection alone cannot distinguish
these cases. They also report objects that are absent (\emph{false positives}), so a single detection does not settle
the target's location either.

Interpreting such observations calls for a world model. Most robot world models reconstruct the scene or predict future
observations~\citep{bar2025navigation,lu2024generative,nie2025wmnav}. Active object search, however, depends on a narrower
quantity: the probability that a view detects the target, given the detector's error rates. Previous methods already
maintain beliefs over the target and model visibility~\citep{zheng2021multi,gao2024probabilistic}, but few predict
detection from probabilistic occlusion and fitted detector error rates together~\citep{zheng2023system,zhou2024aspire,zhou2025respire}.
To our knowledge, no controlled study has isolated which of these components the decisions
use~\citep{yokoyama2024vlfm,liu2026aecnav,zhou2025beliefmapnav}. We therefore ask which world-model representations active
object search requires when the detector is imperfect.

We make two contributions. First, we present Bayesian Active Miss-aware Object Search (BAMOS), a belief-based world model
that predicts detection rather than the scene (Section~\ref{sec:method}). It maintains a target belief over the target's
location and size and an occupancy belief used only to predict visibility, and it trains no network. Second,
the separable components enable a controlled study with an oracle and two open-vocabulary detectors in
Isaac Sim~\citep{nvidia2026isaac} and in Habitat Synthetic Scenes Dataset (HSSD) houses~\citep{khanna2024habitat} in
Habitat-Sim~\citep{savva2019habitat}. The study replaces the occupancy belief with the true geometry, ablates each
component, crosses view and stop rules, and injects detector errors.

\paragraph{Related work.} \emph{Probabilistic object search} updates a belief after non-detections, with a hand-set
detection probability and binary occlusion~\citep{zheng2021multi,zheng2023system}, or with a point
target~\citep{gao2024probabilistic,zhou2024aspire}. BAMOS instead fits the detector's error rates and treats the target's
size as uncertain. \emph{Open-vocabulary object navigation} uses real detectors~\citep{yokoyama2024vlfm,busch2025one} and
lowers map values where nothing was detected, with heuristic weights~\citep{zhou2025beliefmapnav,liu2026aecnav}. BAMOS
replaces these weights with a likelihood. \emph{World models} reconstruct the
scene~\citep{hornung2013octomap,newcombe2011kinectfusion,mildenhall2020nerf,kerbl2023gaussian} or predict
observations~\citep{ha2018world,hafner2023mastering,assran2025vjepa2}, although a model need only be accurate in the
quantities a decision depends on~\citep{grimm2020value,yu2026how}. BAMOS is a case study of this principle
(Appendix~\ref{app:related}).

\section{BAMOS}
\label{sec:method}

\begin{figure}[t]
  \centering
  \includegraphics[width=0.72\linewidth]{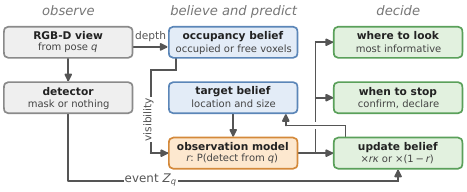}
  \caption{\textbf{One step of BAMOS.} The observation model links both beliefs to the view-selection and stopping
  decisions.}
  \label{fig:arch}
\end{figure}

\paragraph{Task.} A single target object lies in an indoor scene. The robot knows its category but not its location or
size. The target's hidden state is $X=(L,S)$: the cell $L$ that holds its centre and its size scale $S$, with values $v$ and $s$.
At each step the robot teleports its RGB-D camera to a reachable pose $q$, takes a frame, and the detector returns a
mask (\DET) or nothing (\NODET). $D_t$ denotes all observations up to step $t$. The episode ends at a declaration
or after $B$ views. We assume a static scene, a known empty layout with unknown furniture, and exact depth and
poses (Appendix~\ref{app:design}).

\paragraph{Overview.} For each hypothesis of the target's location and size, BAMOS predicts the probability that a view
detects the target (Figure~\ref{fig:arch}). A \emph{target belief} gives the probability that the target is at a
location. An \emph{occupancy belief} over voxels determines how likely a view is to see that location. An
\emph{observation model} combines this visibility with the projected area, the image border and the detector's error
rates. Information gain selects views, and a confirming view precedes each declaration. Three occupancy and six detector
constants are fitted on calibration frames (Appendix~\ref{app:fit}).

\paragraph{Occupancy belief.} The occupancy belief $b_t(g)=P(\text{voxel $g$ occupied}\mid D_t)$ starts at a prior
$p_0$ and is used only to predict visibility. A depth ray that enters a voxel ends there with probability
$q_{\mathrm{occ}}$ if the voxel is occupied and $q_{\mathrm{free}}$ if it is free. If voxel $g$ has been hit
$n^{\mathrm{hit}}_t(g)$ times and passed $n^{\mathrm{pass}}_t(g)$ times, Bayes' rule in odds form gives
\begin{equation}
\operatorname{logit} b_t(g)=\operatorname{logit}p_0+n^{\mathrm{hit}}_t(g)\,\log\frac{q_{\mathrm{occ}}}{q_{\mathrm{free}}}
+n^{\mathrm{pass}}_t(g)\,\log\frac{1-q_{\mathrm{occ}}}{1-q_{\mathrm{free}}}.
\label{eq:occ}
\end{equation}
Hits raise these odds and passes lower them. Unobserved voxels stay at $p_0$ and are partially opaque.

\paragraph{Observation model.} The probability that a view from $q$ detects a target at $(v,s)$ is
\begin{equation}
r_{t,q}(v,s)=\eta_f+(1-\eta_f)\,\eta_t\,T_{t,q}(v,s)\,h\bigl(\phi_q(v,s)\,a_q(v,s)\bigr).
\label{eq:r}
\end{equation}
A detection is either a false positive, at rate $\eta_f$, or a true one. A true detection requires the detector to
fire ($\eta_t$ in plain view), an unoccluded target ($T$), and enough image area ($h$). $T$ is the
probability, under the occupancy belief, that a ray reaches the hypothesis. $h$ is a soft threshold on the visible area,
the expected area $a_q$ times the share $\phi_q$ inside the frame. It rises from 0 to 1 around a fitted area $a_0$, so a
small or distant target is rarely detected. The target belief therefore includes size.

\paragraph{Target belief.} The target belief $\pi_t(v,s)=P(L=v,S=s\mid D_t)$ covers cells on a lattice and a few size
scales, with a uniform prior. A frame from pose $q$ with outcome $o$ updates it by Bayes' rule with likelihood $f$.
Here $\mathbf c$ is the centroid of a \DET's back-projected mask:
\begin{equation}
\pi_{t+1}(v,s)\propto\pi_t(v,s)\,f(o\mid v,s),\qquad
f(o\mid v,s)=\begin{cases} r_{t,q}(v,s)\,\kappa(\mathbf c\mid v,s), & o=(\DET,\mathbf c),\\ 1-r_{t,q}(v,s), & o=\NODET.\end{cases}
\label{eq:update}
\end{equation}
The \emph{mask position} term $\kappa$ mixes a uniform density for false positives
with a Gaussian near the cell centre for true detections (Appendix~\ref{app:model}). In the \emph{non-detection update},
hypotheses the view should have revealed lose most of their probability, while hidden or out-of-frame
ones, with $r_{t,q}(v,s)$ near $\eta_f$, keep most of it. A non-detection thus shifts belief to unobserved regions.

\paragraph{View selection and stopping.} The next view is the unvisited pose $q$ whose detection event $Z_q$ is
most informative about $X$, given the observations so far:
\begin{equation}
U_t(q)=I(Z_q;X\mid D_t)=H_b\Bigl(\textstyle\sum_{v,s}\pi_t(v,s)\,r_{t,q}(v,s)\Bigr)-\textstyle\sum_{v,s}\pi_t(v,s)\,H_b\bigl(r_{t,q}(v,s)\bigr),
\label{eq:ig}
\end{equation}
with $H_b$ the binary entropy. When a \DET\ raises a cell $\hat v$ to probability 0.5, the robot takes a
\emph{confirming view}, which best observes $\hat v$. It declares $\hat v$ if this view detects the target again
and the target lies within $\tau$ of $\hat v$ with probability at least 0.5 (the \emph{belief stop}); otherwise the
search continues.

\section{Experiments}

\subsection{Setup}
\label{sec:setup}

\paragraph{Environments.} In Isaac Sim, 345 test episodes each place a 0.1--0.3\,m cube or sphere behind an occluder, in
four room types. A declaration succeeds within $\tau=0.2$\,m of the target, and the budget is $B=60$ views. In HSSD houses,
1,248 ObjectNav validation episodes in 42 houses each ask for any instance of a category, with $\tau=0.5$\,m and $B=200$
views (Appendix~\ref{app:env}).

\paragraph{Detectors.} The \emph{oracle} reports the target's mask whenever a few pixels are visible. SAM~3~\citep{carion2025sam3}
and Falcon-Perception~\citep{bevli2026falcon} take a text query, and both miss visible targets. SAM~3 returns confidence scores, whose threshold can be
raised to trade more false negatives for fewer false positives.
Falcon-Perception, by contrast, returns no scores to filter its masks, so it reports more false positives.

\paragraph{Comparisons.} We compare BAMOS with four groups of variants by success@$B$, the fraction of episodes declared
correctly within $B$ views. Three groups vary the world model, and one varies the detector. First,
\emph{BAMOS with true geometry} replaces the occupancy belief with a perfect scene reconstruction. Second, ablations
remove the target belief, the occupancy belief, the non-detection update or the mask position ($\kappa\equiv1$). Third,
we cross view rules with stop rules. Fourth, we
inject false negatives or false positives into the oracle at per-frame rates of 0.1--0.3 and refit the constants.

Note that without a target belief, information gain and the \emph{belief stop} do not apply. The robot then picks the pose newly
viewing the most $T$-weighted cells (\emph{visibility-weighted coverage}) and stops at the first detection (\emph{first-mask stop}). The crossed comparison adds unweighted
\emph{field-of-view coverage}.

\begin{table}[t]
\centering\footnotesize
\setlength{\tabcolsep}{4pt}
\begin{tabular}{lcccccc}\toprule
 & \multicolumn{3}{c}{Isaac Sim} & \multicolumn{3}{c}{HSSD houses}\\
\cmidrule(lr){2-4}\cmidrule(lr){5-7}
 & Oracle & SAM 3 & \begin{tabular}[b]{@{}c@{}}Falcon-\\Perception\end{tabular} & Oracle & SAM 3 & \begin{tabular}[b]{@{}c@{}}Falcon-\\Perception\end{tabular}\\\midrule
BAMOS & \textbf{0.977} & \textbf{0.913} & \textbf{0.884} & \underline{0.994} & \underline{0.825} & 0.426\\
BAMOS with true geometry & \underline{0.971} & \underline{0.893} & \underline{0.872} & \textbf{0.995} & \textbf{0.869} & \textbf{0.498}\\\midrule
No target belief & 0.945 & 0.846 & 0.383 & \underline{0.994} & 0.788 & 0.269\\
No occupancy belief & 0.968 & 0.884 & 0.843 & 0.925 & 0.776 & \underline{0.478}\\
No non-detection update & 0.916 & 0.664 & 0.690 & 0.639 & 0.503 & 0.303\\
No mask position & 0.606 & 0.597 & 0.557 & 0.887 & 0.589 & 0.311\\\bottomrule
\end{tabular}
\caption{\textbf{Contribution of each component.} Success@$B$. Best in bold, second underlined.}
\label{tab:main}

\vspace{3pt}
\setlength{\tabcolsep}{4pt}
\begin{tabular}{llccc}\toprule
Views & Stop & Oracle & SAM 3 & Falcon-Perception\\\midrule
Information gain (BAMOS) & belief & \textbf{0.977} & \textbf{0.913} & \textbf{0.884}\\
Visibility-weighted coverage & belief & \underline{0.974} & 0.846 & \underline{0.788}\\
Field-of-view coverage & belief & 0.832 & 0.490 & 0.574\\\midrule
Information gain & first mask & 0.936 & \underline{0.881} & 0.438\\
Visibility-weighted coverage & first mask & 0.945 & 0.846 & 0.383\\
Field-of-view coverage & first mask & 0.878 & 0.612 & 0.423\\\bottomrule
\end{tabular}
\caption{\textbf{View and stop rules.} Success@$B$ in Isaac Sim. Best in bold, second underlined.}
\label{tab:cross}

\vspace{3pt}
\setlength{\tabcolsep}{4pt}
\begin{tabular}{lcccccccc}\toprule
 & \multicolumn{4}{c}{Isaac Sim} & \multicolumn{4}{c}{HSSD houses}\\
\cmidrule(lr){2-5}\cmidrule(lr){6-9}
Per-frame rate & 0.0 & 0.1 & 0.2 & 0.3 & 0.0 & 0.1 & 0.2 & 0.3\\\midrule
False negatives & \multirow{2}{*}{0.977} & 0.974 & 0.974 & 0.959 & \multirow{2}{*}{0.994} & 0.993 & 0.994 & 0.990\\
False positives & & 0.962 & 0.910 & 0.823 & & 0.932 & 0.769 & 0.592\\\bottomrule
\end{tabular}
\caption{\textbf{Effect of injected detector errors.} Success@$B$ with errors injected into the oracle.}
\label{tab:inject}
\end{table}

\subsection{Results}
\label{sec:results}

\paragraph{Can the occupancy belief substitute for the true geometry?} In Isaac Sim, BAMOS matches or slightly exceeds
its true-geometry variant with every detector (Table~\ref{tab:main}). In HSSD houses it trails by 0.001, 0.044 and 0.072 with the oracle,
SAM~3 and Falcon-Perception. This gap likely reflects occupancy errors, more common in cluttered
multi-room HSSD houses with a shorter depth range.

\paragraph{Which components does the search rely on?} Removing a component that the decisions use should lower success
(Table~\ref{tab:main}, Figure~\ref{fig:anytime}). The loss from removing the target belief grows with detector error, from the
oracle through SAM~3 to Falcon-Perception. Removing the non-detection update or the mask
position lowers success in all six settings. Without the mask position, a detection raises every cell in view, so a
wrong cell is more often declared. Removing the occupancy belief costs less than removing either update in every
setting. With Falcon-Perception in HSSD houses, it raises success, suggesting that the occupancy errors noted earlier
can outweigh its benefit.

\paragraph{How do view selection and stopping interact?} Removing the target belief changes both decisions, so
Table~\ref{tab:cross} separates them in Isaac Sim (Appendix~\ref{app:results}). With the belief stop, information gain matches or exceeds both
coverage rules, and weighting cells by $T$ raises coverage success by 0.142--0.356. The view rule matters most with
SAM~3, which needs a target to appear large (highest fitted $a_0$) and so benefits most from well-chosen views. The
stop rule matters most with Falcon-Perception, where a first-mask stop declares its frequent false positives, halving
success under information gain.

\paragraph{How do false negatives and false positives affect the search?} Injected into the oracle (Table~\ref{tab:inject}), false negatives at rates
up to 0.3 reduce success by at most 0.018. False positives reduce it by up to 0.154 in Isaac Sim and 0.402 in HSSD houses. This is likely because an injected false positive usually marks a real object, which the
confirming view can mark again, satisfying the belief stop.

\section{Conclusion}

We asked what world model active object search needs under imperfect detection. BAMOS's success came mainly from how the target belief interprets detections and non-detections, a contribution that
grew as the detector became less reliable. Scene geometry entered only through visibility, and an occupancy belief came close
to the true geometry in success. Within the evaluated scope, a world model can thus centre on
decision-relevant observations, judged by its effect on the decision.

\clearpage

\bibliography{references}

\appendix

\section{Model details}
\label{app:model}

This section gives the implementation of Section~\ref{sec:method}, in the same order.

\subsection{Occupancy belief}

The hit and pass counts of Equation~\eqref{eq:occ} include every voxel that a depth frame observes. Three design choices define
what ``observes'' means.

\paragraph{Rays.} We cast a ray from every second pixel (every fourth in HSSD houses). A ray ends at its measured depth, which is a hit in the voxel
there, and passes every voxel before it. A ray without a valid depth passes all voxels up to
$z_{\max}$, 15\,m in Isaac Sim and 5\,m in HSSD houses. The HSSD sensor clips depths nearer than 0.5\,m. Such a
ray counts as a hit at 0.5\,m and passes nothing. Each voxel counts at most once per frame, and a hit takes precedence over a pass.
Frames are therefore independent given the voxel's state.

\paragraph{Mask pixels.} Inside a reported mask, ray ends are not counted as hits. Otherwise, the target's own surface
would be mistaken for an occluder in front of it. A false positive's mask may cover a real object, whose surface is then
not counted either.

\paragraph{Order.} A frame first updates the target belief with Equation~\eqref{eq:update} and only then the occupancy
belief. The detection probability $r$ thus always uses the occupancy belief from before the frame.

\subsection{Observation model}

The detection probability of Equation~\eqref{eq:r} has three geometric factors: the visibility $T$, the expected image
area $a$ and the in-image share $\phi$. The soft threshold $h$ turns their product into a detection rate.

\paragraph{Visibility.} A ray that enters voxel $g$ passes through it without terminating with probability
\begin{equation}
p^{\mathrm{pass}}_t(g)=1-b_t(g)\,q_{\mathrm{occ}}-\bigl(1-b_t(g)\bigr)\,q_{\mathrm{free}}.
\end{equation}
Since voxels are independent, a ray from the camera position $\mathbf p_q$ reaches a point $\mathbf y$ unoccluded with a probability equal to
the product of these probabilities along the ray. The visibility of a hypothesis averages this product over nine points
$\mathbf y_1,\dots,\mathbf y_9$ of its front face (Equation~\eqref{eq:tbar}). The front face is a square of side $s$ through
$\mathbf x_v+\tfrac s2\mathbf n_q$, facing the camera, and the points form a $3\times3$ grid on it:
\begin{equation}
T_{t,q}(v,s)=\frac19\sum_{k=1}^{9}\ \prod_{g\in\mathrm{ray}(\mathbf p_q,\mathbf y_k)}p^{\mathrm{pass}}_t(g).
\label{eq:tbar}
\end{equation}
Two segments of each ray are excluded: its last $\delta=0.1$\,m and the camera's own voxel. The margin prevents the supporting surface,
and the target's own surface when the cell centre is offset from the target's centre, from counting as occlusion. As a consequence,
an occluder within $\delta$ of the target is ignored. $T$ depends only on the camera position, so all
orientations at a position share it.

\paragraph{Expected image area.} A body of size scale $s$ whose centre lies at depth $z$ along the optical axis covers,
on average, the image fraction
\begin{equation}
a_q(v,s)=\frac{f_xf_y\,s^2}{N_xN_y\,\max\{z,z_{\min}\}^2},
\end{equation}
with focal lengths $f_x,f_y$ in pixels, an image of $N_x\times N_y$ pixels and $z_{\min}$ the camera's near limit (0.1\,m; HSSD 0.5\,m). It decreases with the
square of the distance and increases with the square of the size.

\paragraph{In-image share.} A target truncated by the image border is only partially visible. We model its silhouette as a
square of the same area, of side $w=\sqrt{f_xf_y}\,s/z$ pixels, centred on the projection $(\bar x,\bar y)$ of
$\mathbf x_v$. The share of this square inside the image is
\begin{align}
\phi_q(v,s)&=\mathbf 1\{z_{\min}\le z\le z_{\max}\}\,\rho(\bar x;N_x)\,\rho(\bar y;N_y),\\
\rho(x;n)&=\Bigl[\frac{\min\{x+w/2,\,n-\tfrac12\}-\max\{x-w/2,\,-\tfrac12\}}{w}\Bigr]_0^1,
\end{align}
where $[\cdot]_0^1$ clips to $[0,1]$. A test on the centre alone would set $r=\eta_f$ for a hypothesis whose centre is
just outside the image although part of its body appears in the image. $\phi$ retains the visible fraction instead.

\paragraph{Soft threshold.} The detector fires on the visible area $\phi_qa_q$ with relative rate $h(\phi_qa_q)$, where
\begin{equation}
h(x)=\frac{x^\gamma}{x^\gamma+a_0^\gamma}.
\end{equation}
It is 0 for a vanishing area, $\tfrac12$ at the midpoint $a_0$, and 1 for a large area. The slope $\gamma$ sets how
sharply it rises.

\subsection{Target belief}

\paragraph{Size scale.} Hypotheses are convex bodies without orientation. Averaged over all directions, a convex body's
projected area is a quarter of its surface area $A$ (Cauchy's formula). We therefore set
\begin{equation}
s=\sqrt{A/4},
\end{equation}
so that $s^2$ is the mean silhouette area. For a sphere of radius $R$, $s=\sqrt\pi R$. For a cube of side $l$,
$s=\sqrt{3/2}\,l$.

\paragraph{Cell average.} $L=v$ means that the target's centre lies anywhere in cell $v$, not exactly at its centre. We
therefore average $r$ and $r\kappa$ over the cell's eight octant centres $\mathbf x_v+\{\pm\Delta/4\}^3$, with $\Delta$ the cell size (0.2\,m in
Isaac Sim and 0.25\,m in HSSD houses). Without
this, a target truncated by the image border could be detected while its cell centre predicts $r=\eta_f$. Fitting uses the same
average.

\paragraph{One non-detection.} Consider five equally likely cells: three in plain view with $r=0.9$, one hidden behind
an occluder with $r=0.05$, and one outside the frame with $r=0.006$. After one \NODET, Equation~\eqref{eq:update} multiplies
each cell by $1-r$. The three cells in plain view keep a tenth of their mass each. The hidden cell ($r=0.05$) and the cell
outside the frame ($r=0.006$) keep almost all of theirs, so after normalisation they hold 0.42 and 0.44 of the belief.

\paragraph{From a mask to a centroid.} We keep one mask: for SAM~3 the highest-scoring mask above the score threshold,
for Falcon-Perception, which provides no scores, the largest. Each mask pixel $(x,y)$ with a valid depth $0<d\le z_{\max}$
gives the camera-frame point
\begin{equation}
\mathbf p=d\,\Bigl[\frac{x-c_x}{f_x},\ \frac{y-c_y}{f_y},\ 1\Bigr]^\top,
\end{equation}
with $d$ the depth along the optical axis and $(c_x,c_y)$ the principal point. The known pose maps $\mathbf p$ to the
world, and the centroid $\mathbf c$ is the mean of these world points. With fewer than three valid pixels, $\kappa\equiv1$.
Because $\mathbf c$ is the centre of the visible surface, not of the body, it lies towards the camera. The offset
$\beta s$ in Equation~\eqref{eq:kappa-main} below captures this, and $\lambda s$ the remaining spread.

\paragraph{Mask position.} The mask position term $\kappa$ of Equation~\eqref{eq:update}, a density over the centroid,
mixes the two kinds of detection:
\begin{equation}
r_{t,q}(v,s)\,\kappa(\mathbf c\mid v,s)=\eta_f\,u+\bigl(r_{t,q}(v,s)-\eta_f\bigr)\,\mathcal N\bigl(\mathbf c;\ \mathbf x_v+\beta s\,\mathbf n_q,\ (\lambda s)^2I\bigr),
\label{eq:kappa-main}
\end{equation}
with $\mathbf n_q$ the unit vector from the cell centre towards the camera. A detection is a
false positive with probability $\eta_f/r$, since false positives make up $\eta_f$ of the detection probability $r$. A
false positive's mask lies anywhere, so its centroid is uniform with density $u=1/|\mathcal W|$ over the search space,
of volume $|\mathcal W|$. Otherwise the detection is true, and its centroid is Gaussian around
$\mathbf m=\mathbf x_v+\beta s\,\mathbf n_q$. Weighting the two densities by these probabilities and multiplying
by $r$ gives Equation~\eqref{eq:kappa-main}. $\kappa$ integrates to 1, so the likelihood $f$ of Equation~\eqref{eq:update}
integrates to $r+(1-r)=1$ over all outcomes. The term $\eta_f u$ acts as a floor: one centroid far from a hypothesis,
such as a false positive's, cannot eliminate that hypothesis.

\paragraph{Several masks (used in HSSD houses).} If the detector returns $M$ masks above the threshold, with centroids
$\mathbf c_1,\dots,\mathbf c_M$, we assume that at most one is the target, each with probability $1/M$. The
others are false positives with uniform centroids. Up to the factor $u^{M-1}$, which every hypothesis shares and the
normalisation of $\pi$ cancels,
\begin{equation}
r_{t,q}(v,s)\,\kappa(\mathbf c_1,\dots,\mathbf c_M\mid v,s)\propto\eta_f\,u+\bigl(r_{t,q}(v,s)-\eta_f\bigr)\,\frac1M\sum_{i=1}^M\mathcal N\bigl(\mathbf c_i;\ \mathbf m,\ (\lambda s)^2I\bigr).
\end{equation}
In Isaac Sim, this did not change success with SAM~3 (several masks in 4 of 1073 detecting views). With Falcon-Perception,
success was 0.872 ($-0.012\,[-0.029,+0.006]$ relative to one mask), so the Isaac Sim runs keep one mask. HSSD houses use all masks, since a house
contains several objects of a category.

\subsection{View selection and stopping}

\paragraph{Information gain.} A \DET\ also carries its centroid, so its full information is
\begin{equation}
I(Z_q,\mathbf C_q;X\mid D_t)=I(Z_q;X\mid D_t)+\bar r_t(q)\,I(\mathbf C_q;X\mid\DET,D_t),
\end{equation}
with $\mathbf C_q$ the centroid at view $q$ as a random variable and $\bar r_t(q)=\sum_{v,s}\pi_t(v,s)\,r_{t,q}(v,s)$ the predicted detection probability. The second term requires the entropy of a mixture of Gaussians and a
uniform, which has no closed form. It would have to be estimated for every candidate pose at every step. With rare false
positives, it is close to a monotone function of $\bar r_t(q)$, and $I(Z_q;X\mid D_t)$ already ranks views the same way when
$\bar r_t(q)\ll\tfrac12$. We therefore use $U_t(q)=I(Z_q;X\mid D_t)$ alone (Equation~\eqref{eq:ig}). With frequent false positives (Falcon-Perception), dropping the
second term may undervalue close views. Ties between utilities within $10^{-12}$ are broken by the shorter path, then by the lower
pose index. In Isaac Sim, choosing a view took a median of 1.0\,s (95th percentile 6.3\,s) on an A100 GPU shared with the renderer.

\paragraph{Confirming view.} When a \DET\ makes the most probable cell $\hat v$ reach $\pi_t(\hat v)\ge0.5$, with $\pi_t(v)=\sum_s\pi_t(v,s)$, the robot
moves to the pose that maximises the expected visible area of $\hat v$'s body,
\begin{equation}
V_t(q)=\textstyle\sum_s\pi_t(s\mid\hat v)\,\phi_q(\hat v,s)\,a_q(\hat v,s)\,T_{t,q}(\hat v,s),
\end{equation}
among the poses that keep the body inside the image at all eight octant points for every size with
$\pi_t(s\mid\hat v)\ge0.05$.

\paragraph{Probability of success.} The declaration succeeds if the target lies within $\tau$ of $\hat v$'s centre. Its
probability under the target belief is
\begin{equation}
P_{\mathrm{succ}}=\textstyle\sum_{v'}\omega(v'-\hat v)\,\pi_t(v'),
\end{equation}
where $\omega(v'-\hat v)$ is the share of cell $v'$ within $\tau$ of $\hat v$'s centre. In Isaac Sim, it is 1 for
$\hat v$ itself, 0.41 for a face neighbour and 0.06 for an edge neighbour.

\paragraph{Declaration.} After the confirming view, the robot declares $\hat v$ if three conditions hold. The confirming
view detects the target again, the most probable cell remains within $\tau$ of $\hat v$, and $P_{\mathrm{succ}}\ge0.5$.
Otherwise, it withdraws the candidate and continues the search. The confirming view counts towards $B$ like any other view.

\subsection{One step}

Each step first lists the reachable, unvisited poses. For every pose and hypothesis, the robot computes $a$, $\phi$ and
$T$ at the eight octant points, and from them $r$. It selects the pose with the highest $U$, or with the highest $V$ if
a confirming view is pending, and stores that pose's $r$. After taking the frame, it updates $\pi$ with
Equation~\eqref{eq:update}, then $b$ with Equation~\eqref{eq:occ}, and checks the stop rule. A frame in which the detector fails to run leaves $\pi$ unchanged, but its depth still updates $b$.

\FloatBarrier
\section{Fitting the constants}
\label{app:fit}

\paragraph{Occupancy constants.} $p_0$, $q_{\mathrm{occ}}$ and $q_{\mathrm{free}}$ describe the scene and the depth
sensor, so all detectors share them. Over all calibration depth frames, a voxel observed $n$ times collects
$n^{\mathrm{hit}}$ hits. If the voxel is occupied with prior probability $p_0$, its hit count follows the two-component
mixture
\begin{equation}
n^{\mathrm{hit}}\sim p_0\,\mathrm{Bin}(n,q_{\mathrm{occ}})+(1-p_0)\,\mathrm{Bin}(n,q_{\mathrm{free}}),
\end{equation}
and the three constants are fitted by maximum likelihood.

\paragraph{Detector constants.} $\eta_f$, $\eta_t$, $\gamma$, $a_0$, $\beta$ and $\lambda$ are fitted jointly by maximising the
likelihood $f$ of Equation~\eqref{eq:update} over the calibration frames. For each frame we evaluate $f$ at the lattice
hypothesis nearest the true target, with the model's own area and with $T$ from the episode's earlier frames. The
occupancy constants thus enter through Equation~\eqref{eq:occ}. The oracle's $a_0$ is fixed at its detection threshold. SAM~3's
score threshold was chosen from $\{0.5,0.6,\dots,0.9\}$ to maximise calibration success.

\paragraph{Fitted values.} Table~\ref{tab:fit} lists them. With false positives injected into the oracle at rates 0.1, 0.2 and 0.3,
$\eta_f$ is 0.052, 0.108 and 0.158 in Isaac Sim, and 0.070, 0.130 and 0.182 in HSSD houses.

\begin{table}[h]
\centering\footnotesize
\setlength{\tabcolsep}{5pt}
\begin{tabular}{lcccccc}\toprule
 & \multicolumn{3}{c}{Isaac Sim} & \multicolumn{3}{c}{HSSD houses}\\
\cmidrule(lr){2-4}\cmidrule(lr){5-7}
 & Oracle & SAM 3 & \begin{tabular}[b]{@{}c@{}}Falcon-\\Perception\end{tabular} & Oracle & SAM 3 & \begin{tabular}[b]{@{}c@{}}Falcon-\\Perception\end{tabular}\\\midrule
$p_0$, $q_{\mathrm{occ}}$, $q_{\mathrm{free}}$ & \multicolumn{3}{c}{0.042, 0.744, 0.00035} & \multicolumn{3}{c}{0.061, 0.805, 0.00076}\\\midrule
$\eta_f$ & 0.0060 & 0.0076 & 0.087 & 0.018 & 0.023 & 0.264\\
$\eta_t$ & 0.999 & 0.999 & 0.999 & 0.999 & 0.895 & 0.999\\
$\gamma$ & 3.60 & 2.91 & 3.19 & 100 & 8.21 & 87.3\\
$a_0$ & $10^{-4}$\,$^\dagger$ & $7.3\times10^{-4}$ & $1.9\times10^{-4}$ & $4.9\times10^{-4}$\,$^\dagger$ & $2.3\times10^{-3}$ & $5.3\times10^{-5}$\\
$\beta$ & 0.33 & 0.16 & $-0.03$ & 0.19 & 0.15 & 0.16\\
$\lambda$ & 0.36 & 0.36 & 0.57 & 0.16 & 0.16 & 0.18\\\midrule
Score threshold & -- & 0.5 & -- & -- & 0.9 & --\\\bottomrule
\end{tabular}
\caption{\textbf{Fitted constants.} The occupancy constants are shared by all detectors in an environment. $^\dagger$: fixed at
the oracle's detection threshold.}
\label{tab:fit}
\end{table}

\FloatBarrier
\section{Experimental design}
\label{app:design}

This section expands the setup of Section~\ref{sec:setup}.

\paragraph{Assumptions.} We make five assumptions. (A1)~The scene is static. (A2)~The robot knows the empty layout,
its walls and floors, but not the objects in it, such as furniture, or the target. (A3)~The robot teleports
between reachable poses, and cost is counted in the number of views. A navigation oracle gives the reachable poses
(Appendix~\ref{app:env}). (A4)~Only the frame at each chosen pose is used, not frames taken while
moving. (A5)~Depth and camera poses are exact within the sensor's range. A3 and A4 remove path following and the frames
taken on the way. Every compared method thus receives the same observations and differs only in what it believes and where
it looks. Floor plans (A2) are often available for warehouses, offices and homes. The target belief has a uniform prior and assumes a
single target, although HSSD episodes accept any of several instances. We compare variants only within each
environment, since the two differ in scenes, targets, success radius and budget (Table~\ref{tab:env}).

\paragraph{Controlled comparisons.} Table~\ref{tab:design} lists what we vary. Removing the target belief changes two
aspects at once: how views are chosen and how the robot stops. We therefore also cross three view rules with two stop
rules (Table~\ref{tab:cross}).

\begin{table}[h]
\centering\footnotesize
\setlength{\tabcolsep}{4pt}
\begin{tabular}{ll}\toprule
Factor & Levels\\\midrule
Detector & oracle; SAM~3; Falcon-Perception\\
Scene information & occupancy belief; true geometry (perfect reconstruction)\\
Representation & full; no occupancy belief; no target belief\\
Mechanism & no non-detection update; no mask position ($\kappa\equiv1$)\\
View rule (Isaac Sim) & information gain; visibility-weighted coverage; field-of-view coverage\\
\quad $\times$ stop rule & belief (confirm, then $P_{\mathrm{succ}}$); first mask\\
Injected errors (oracle) & false negatives, false positives at 0.1, 0.2, 0.3\\\midrule
Fixed & motion (teleport, same candidate poses); uniform prior;\\
 & budget ($B=60$; HSSD: 200); test episodes (HSSD: validation)\\\bottomrule
\end{tabular}
\caption{\textbf{Varied and fixed factors.} Variants are paired by
episode. HSSD houses run all factors except the crossed view and stop rules.}
\label{tab:design}
\end{table}

\paragraph{Metrics.} We report success@$B$, the fraction of episodes with a correct declaration within $B$ views. Each variant runs every episode once with seed 0. Constants are fitted on separate calibration
episodes. The design, including $B=60$, was developed on separate development splits.

\FloatBarrier
\section{Additional results}
\label{app:results}

\begin{figure}[h]
  \centering
  \includegraphics[width=\linewidth]{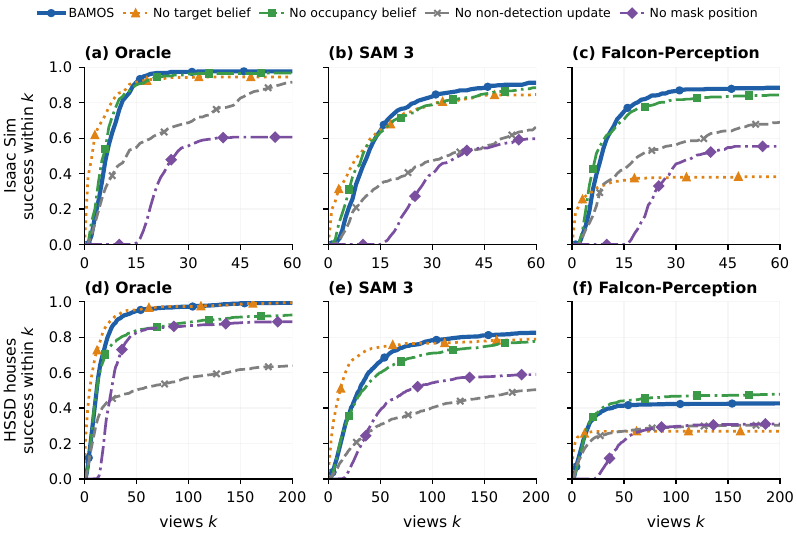}
  \caption{\textbf{Success within $k$ views.} Top: Isaac Sim ($B=60$). Bottom: HSSD houses ($B=200$). The value at $k=B$ is the success@$B$ of Table~\ref{tab:main}.}
  \label{fig:anytime}
\end{figure}

This section expands the results of Section~\ref{sec:results}. Figure~\ref{fig:anytime} gives the success within $k$ views.

\paragraph{True geometry.} The true geometry uses the occupancy grid (0.05\,m in Isaac Sim, 0.1\,m in HSSD houses).
Every voxel that holds part of an object other than a target is opaque, with a pass probability of $10^{-6}$. Depth frames do not change it. $T$
is computed as for the occupancy belief (Appendix~\ref{app:model}). The constants are not refitted: they were fitted
with the occupancy belief's visibility.

\paragraph{Crossed view and stop rules.} The crossed comparison in Isaac Sim separates the two uses of the belief
(Table~\ref{tab:cross}). The two open-vocabulary detectors differ. With SAM~3, whose fitted area threshold $a_0$ is the highest,
the view rule has the largest effect. With Falcon-Perception, whose $\eta_f$ is the highest, the stop rule has the largest
effect (Section~\ref{sec:results}). With the oracle, both effects are small, except that field-of-view coverage
reduces success by up to 0.145.

\paragraph{The oracle without a target belief.} With the oracle, removing the target belief changes success by at most
0.032. In Isaac Sim, its wrong first-mask declarations fall just outside the 0.2\,m success radius. In HSSD houses, where
the radius is 0.5\,m, none occur.

\paragraph{The budget.} Because BAMOS takes a confirming view before each declaration, it requires more views than a
first-mask stop. With the oracle and SAM~3, the variant without a target belief therefore succeeds more often at small
budgets in both environments. In Isaac Sim, BAMOS exceeds it within about 20 views. In HSSD houses, the variant stays
ahead for longer, and BAMOS matches it with the oracle and exceeds it with SAM~3 at larger budgets. With Falcon-Perception, the variant declares early on false positives in both
environments.

\paragraph{The mask position.} Without the mask position, no episode succeeds within fewer than 10 views in either
environment. A detection then raises every cell in view, so the belief concentrates on a single cell only after further
views, with or without detections, have narrowed down the candidate cells.

\paragraph{False positives.} The observation model treats false positives as independent across frames, with uniform
centroids. A false positive that recurs at the same location in the confirming view violates this assumption and
satisfies the belief stop. The sensitivity to false positives thus reflects the current stop rule and false-positive
model rather than the use of a belief. Whether a stricter confirmation or a model of recurring false positives would
reduce it remains an open question.

\FloatBarrier
\section{Extended related work}
\label{app:related}

\paragraph{Probabilistic object search.} Search theory treats an unsuccessful look as evidence~\citep{chung2012analysis}.
3D-MOS and GenMOS keep a belief over an octree of voxels and treat the object
as one voxel~\citep{zheng2021multi,zheng2023system}. A visible target voxel yields its label with a hand-set
probability, and a hidden one yields no information. Occlusion comes from ray tracing a deterministic octree. 3D-MOS also
varies the sensor noise, a controlled detector study outside object navigation. BPOD, ASPIRe and ReSPIRe multiply the
probability of being in the field of view by the probability of an unblocked line of sight to a point
target~\citep{gao2024probabilistic,zhou2024aspire,zhou2025respire}. BAMOS's $a$, $\phi$ and $T$ extend this to
a finite body whose size is itself uncertain. A recent partially observable Markov decision process (POMDP) maintains a belief over which observed object is the
target~\citep{chen2026pomdp}. None of the methods above fits its observation constants from data or models where a false
detection's mask falls. BAMOS fits both error types per detector: $\eta_t$ and $a_0$ for false negatives, and $\eta_f$
with the mask position term $\kappa$ for false positives.

\paragraph{Open-vocabulary object navigation.} VLFM builds a value map from a vision-language model and navigates with a
learned point-goal policy~\citep{yokoyama2024vlfm}. One Map maintains a single feature map and plans with
A*~\citep{busch2025one}. Uncertainty-aware search lowers map values according to prompt uncertainty~\citep{bajpai2025uncertainty}. WMNav lets a vision-language model imagine and choose the next move~\citep{nie2025wmnav}.
BeliefMapNav and AECNav lower a voxel or cluster score after an observation without a detection, weighted by distance, image position or the
share in the field of view~\citep{zhou2025beliefmapnav,liu2026aecnav}. BeliefMapNav's weights are products or minima, not
normalised likelihoods, and AECNav maintains log-odds per cluster with a fixed empirical false-negative rate. Neither models occlusion, size or false positives. Detector confidence alone does not
separate occlusion from absence~\citep{he2026detector}, which motivates BAMOS to predict detection from geometry rather
than from scores. BeliefMapNav's semantic prior could replace BAMOS's uniform prior.

\paragraph{World models.} Occupancy grids keep a per-voxel occupancy probability~\citep{elfes1989using,hornung2013octomap},
as BAMOS's occupancy belief does. Its ray-based visibility follows active reconstruction~\citep{isler2016information}.
Learned world models predict latent states or observations~\citep{ha2018world,hafner2023mastering,assran2025vjepa2,bar2025navigation},
although a model need only be accurate in the quantities the decision depends on~\citep{grimm2020value,grimm2021proper}.
GenEx imagines observations to update a belief for exploration~\citep{lu2024generative}. To predict whether a view would
detect a target not yet seen, a learned world model must still hypothesise where it is, which is what the target belief represents.
A POMDP belief over the full hidden state is a sufficient statistic~\citep{kaelbling1998planning}. In our experiments,
a simpler, belief-based model that predicts detection supported the search. Its target belief compensated for many of
the detector errors, and an occupancy belief used only for visibility matched the true geometry in Isaac Sim.

\FloatBarrier
\section{Environments and implementation}
\label{app:env}

Table~\ref{tab:env} compares the two environments. The model and its update, view and stop rules are the same in both.

\begin{table}[h]
\centering\footnotesize
\setlength{\tabcolsep}{4pt}
\begin{tabular}{lll}\toprule
 & Isaac Sim & HSSD houses\\\midrule
Episodes & 345 test, four room types & 1,248 validation, 42 houses, six categories\\
Target & one cube or sphere & a median of three instances of a category\\
Target cells, success & 0.2\,m; within 0.2\,m & 0.25\,m; within 0.5\,m of a goal box\\
Sizes & 0.1, 0.2, 0.3\,m & 10th, 50th, 90th percentile in training houses\\
Occupancy voxels & 0.05\,m & 0.1\,m\\
Camera & $320\times240$, $70^\circ$, 1.5\,m high & $640\times480$, $79^\circ$, 0.88\,m high\\
Depth range & 0.1--15\,m & 0.5--5\,m\\
Poses (0.5\,m grid) & 8 yaws, 2 pitches & 12 yaws, 2 pitches\\
Next pose & reachable, unvisited & unvisited, within 4\,m, on seen floor\\
Known geometry & room bounds & walls, floors, stairs, doors, windows, railings\\
Masks (open-vocabulary) & one & every mask above the threshold\\
SAM~3 score threshold & 0.5 & 0.9 (chosen on training houses)\\
Oracle & 1\% of the target and $\sim$8 pixels & 150 pixels with valid depth\\
Budget & $B=60$ & $B=200$\\
Constants fitted on & calibration episodes & 214 episodes in training houses\\\bottomrule
\end{tabular}
\caption{\textbf{The two environments.}}
\label{tab:env}
\end{table}

\paragraph{Scenes and targets.} In Isaac Sim, targets are placed on or inside furniture, with their centres 0.1\,m off
the candidate lattice. No robot pose lies within 1.2\,m of the target. Episodes are kept only if some pose shows at
least half of the target to the oracle. Of the 345 test episodes, 100 are in the large warehouse. In HSSD houses, each
episode accepts any instance of its category, and the size hypotheses are percentiles of that category's goal sizes in
the training houses. Nine episodes in two houses start on a roof slab, a defect in the benchmark. We retain them.

\paragraph{Poses.} In Isaac Sim, reachability comes from a 0.1\,m floor map of the true obstacle footprints, inflated by
the robot's clearance. Every pose connected to the start is a candidate from the first step. The search model never
reads this map, but the candidate set reveals where the floor is free, so every compared method receives the same scene
information. In HSSD houses, the camera follows the benchmark, and candidate poses are restricted to floor seen so far.
In both environments, path length only breaks ties.

\paragraph{Detectors and runs.} SAM~3 and Falcon-Perception are queried with a fixed phrase per target category. Each
variant runs every episode once, with planner and sensor seed 0, and variants are paired by episode. Code will be
released.

\FloatBarrier
\section{Qualitative example: narrowing the target belief with SAM~3}
\label{app:trace}

\begin{figure}[h]
  \centering
  \includegraphics[width=\linewidth]{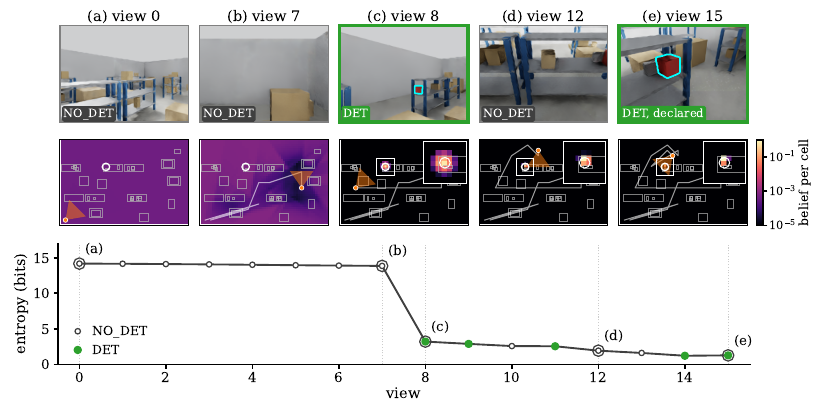}
  \caption{\textbf{One search with SAM~3 in Isaac Sim.} Top: the frame at five views (green border: \DET; cyan: the
  reported mask). Middle: the target belief from above, summed over heights and sizes, on one log scale for all views. Grey
  outlines: furniture. White ring: the target. Orange wedge: the camera's field of view. White line: the path. Once the
  belief has concentrated, the inset enlarges a 1.6\,m window around the target, and its ring is the 0.2\,m success
  radius. Bottom: the entropy of the belief over cells after each view, with the five views ringed.}
  \label{fig:trace}
\end{figure}

Figure~\ref{fig:trace} follows one Isaac Sim test episode with SAM~3, in which the target is a cube of size 0.3\,m on a shelf
of the large warehouse. Before the first detection, each non-detection lowers the belief in the cells the camera has seen
(b). Because the room is large, the entropy falls only slowly, from 14.2 to 13.8 bits over seven views. The first
detection (c) concentrates the belief around the mask's centroid, and the entropy drops to 3.2 bits. Later views narrow
it further. At view 12 (d), the target is cut off at the image border and SAM~3 misses it, but the belief stays near the
target. At view 14, a detection raises the most probable cell above 0.5. The confirming view (e) detects the target
again, and the search declares a cell 0.19\,m from the target's centre.

\end{document}